\documentclass[sn-mathphys-num]{sn-jnl}% Math and Physical Sciences Numbered Reference Style 
\usepackage{graphicx}%
\usepackage{multirow}%
\usepackage{amsmath,amssymb,amsfonts}%
\usepackage{amsthm}%
\usepackage{mathrsfs}%
\usepackage[title]{appendix}%
\usepackage{xcolor}%
\usepackage{textcomp}%
\usepackage{manyfoot}%
\usepackage{booktabs}%
\usepackage{algorithm}%
\usepackage{algorithmicx}%
\usepackage{algpseudocode}%
\usepackage{listings}%
\usepackage{caption}
\theoremstyle{thmstyleone}%
\theoremstyle{thmstyletwo}%

\theoremstyle{thmstylethree}%

\begin{document}

\title[Article Title]{AURA-CVC: Autonomous Ultrasound-guided Robotic Assistance for Central Venous Catheterization}

%%=============================================================%%
%% GivenName	-> \fnm{Joergen W.}
%% Particle	-> \spfx{van der} -> surname prefix
%% FamilyName	-> \sur{Ploeg}
%% Suffix	-> \sfx{IV}
%% \author*[1,2]{\fnm{Joergen W.} \spfx{van der} \sur{Ploeg} 
%%  \sfx{IV}}\email{iauthor@gmail.com}
%%=============================================================%%
% \author{Deepak Raina$^1$, Lidia Al-Zogbi$^2$, Thorsten Fleiter$^3$,  Muyinatu A. Lediju Bell$^{1,4}$, Vinciya Pandian$^5$, \\Axel Krieger$^{1,2}$
% \thanks{This work is supported by the National Institutes of Health under award numbers R01EB020667A1 and R18HS029124, National Science Foundation under CAREER identification number 2144348, and Malone Postdoctoral Fellowship.}
% \thanks{$^{1}$Deepak Raina, Muyinaty A. Lediju Bell, and Axel Krieger are with the Malone Center for Engineering in Healthcare, Johns Hopkins University, Baltimore, MD, USA. Corresponding author email: \tt\small draina1@jh.edu.}
% \thanks{$^{2}$Lidia Al-Zogbi and Axel Krieger are with the Department of Mechanical Engineering, Laboratory of Computational Sensing and Robotics, Johns Hopkins University, Baltimore, MD, USA.}
% \thanks{$^{3}$Thorsten Fleiter is with the R. Cowley Shock Trauma Center, University of Maryland, Baltimore, MD, USA.}
% \thanks{$^{4}$Muyinaty A. Lediju Bell is with the Departments of Electrical Engineering, Biomedical Engineering, Computer Science, and Oncology, Johns Hopkins University, Baltimore, MD, USA.}
% \thanks{$^{5}$Vinciya Pandian is with the School of Nursing, Pennsylvania State University, University Park, PA, USA.}
\author[1]{\fnm{Deepak} 
\sur{Raina}}
% \email{draina1@jhu.edu}
\author[1]{\fnm{Lidia} \sur{Al-Zogbi}}
% \equalcont{These authors contributed equally to this work.}

\author[2]{\fnm{Brian} \sur{Teixeira}}
\author[2]{\fnm{Vivek} \sur{Singh}}
\author[2]{\fnm{Ankur} \sur{Kapoor}}
% \equalcont{These authors contributed equally to this work.}
\author[3]{\fnm{Thorsten} \sur{Fleiter}}
\author[1]{\fnm{Muyinatu} \sur{A. Lediju Bell}}
\equalcont{These authors advised this work equally.}
\author[4]{\fnm{Vinciya} \sur{Pandian}}
\equalcont{These authors advised this work equally.}
\author*[1]{\fnm{Axel} \sur{Krieger}}\email{axel@jhu.edu}
\equalcont{These authors advised this work equally.}
% \affil[1]{\orgdiv{Malone Center for Engineering in Healthcare}}
% \affil[2]{\orgdiv{Department of Mechanical Engineering}}
% \affil[5]{\orgdiv{Departments of Electrical Engineering,}
% \orgname{Johns Hopkins University}, \orgaddress{\street{3400 N Charles St}, \city{Baltimore}, \postcode{21218}, \state{MD}, \country{USA}}}
% \affil[3]{\orgdiv{Medical Imaging Technologies}, \orgname{Siemens Healthineers}, \orgaddress{\street{755 College Rd E}, \city{Princeton}, \postcode{08540}, \state{NJ}, \country{USA}}}
% % \affil[4]{\orgdiv{R. Cowley Shock Trauma Center, Department of Diagnostic Radiology, School of Medicine}, \orgname{University of Maryland}, \orgaddress{\street{22 S Greene St}, \city{Baltimore}, \postcode{21201}, \state{MD}, \country{USA}}}
% \affil[4]{\orgdiv{Department of Diagnostic Radiology}, \orgname{University of Maryland}, \orgaddress{\street{22 S Greene St}, \city{Baltimore}, \postcode{21201}, \state{MD}, \country{USA}}}
% \affil[6]{\orgdiv{Ross and Carol Nese College of Nursing}, \orgname{Penn State University}, \orgaddress{\street{Nursing Sciences Building}, \city{University Park}, \postcode{16802}, \state{PA}, \country{USA}}}
\affil[1]{\orgname{Johns Hopkins University}, \orgaddress{\city{Baltimore}, \state{MD}, \country{USA}}}
\affil[2]{\orgname{Siemens Healthineers}, \orgaddress{\city{Princeton}, \state{NJ}, \country{USA}}}
% \affil[4]{\orgdiv{R. Cowley Shock Trauma Center, Department of Diagnostic Radiology, School of Medicine}, \orgname{University of Maryland}, \orgaddress{\street{22 S Greene St}, \city{Baltimore}, \postcode{21201}, \state{MD}, \country{USA}}}
\affil[3]{\orgname{University of Maryland}, \orgaddress{\city{Baltimore}, \state{MD},  \country{USA}}}
\affil[4]{\orgname{Penn State University}, \orgaddress{\city{University Park}, \state{PA}, \country{USA}}}
%%==================================%%
%% Abstract guidelines %%
%%==================================%%
% Please provide a structured abstract of 150 to 250 words, which should be divided into the following sections:
% - Purpose (stating the main purposes and research question)
% - Methods
% - Results
% - Conclusion
\abstract{\textit{Purpose:} Central venous catheterization (CVC) is a critical medical procedure for vascular access, hemodynamic monitoring, and life-saving interventions. Its success remains challenging due to the need for continuous ultrasound-guided visualization of a target vessel and approaching needle, which is further complicated by anatomical variability and operator dependency. 
Errors in needle placement can lead to life-threatening complications. While robotic systems offer a potential solution, achieving full autonomy remains challenging. In this work, we propose an end-to-end robotic-ultrasound-guided CVC pipeline, from scan initialization to needle insertion.
% Robotic systems offer a promising solution to automate ultrasound scanning, anatomical identification, and needle guidance, addressing challenges such as procedural variability and visualization constraints.
% However, full autonomy, spanning from scan initialization to needle insertion, remains largely unexplored due to the complexities of real-time anatomical interpretation and robust motion planning of the robot. 

\textit{Methods:} We introduce a deep-learning model to identify clinically relevant anatomical landmarks from a depth image of the patient's neck, obtained using RGB-D camera, to autonomously define the scanning region and paths. Then, a robot motion planning framework is proposed to scan, segment, reconstruct, and localize vessels (veins and arteries), followed by the identification of the optimal insertion zone. Finally, a needle guidance module plans the insertion under ultrasound guidance with operator's feedback. This pipeline was validated on a high-fidelity commercial phantom across 10 simulated clinical scenarios. 

\textit{Results:} The proposed pipeline achieved 10 out of 10 successful needle placements on the first attempt.  Vessels were reconstructed with a mean error of 2.15 \textit{mm}, and autonomous needle insertion was performed with an error less than or close to 1 \textit{mm}. 

\textit{Conclusion:} To our knowledge, this is the first robotic CVC system demonstrated on a high-fidelity phantom with integrated planning, scanning, and insertion. Experimental results show its potential for clinical translation.}
% To our knowledge, this is the first demonstration of a robotic system for CVC placement in a high-fidelity phantom, integrating autonomous intra-operative planning, robotic scanning, and  needle insertion. Experimental results show its potential for clinical translation.}

\keywords{Robotic ultrasound, ultrasound-guided interventions, central venous catheterization, motion-planning}

%%\pacs[JEL Classification]{D8, H51}

%%\pacs[MSC Classification]{35A01, 65L10, 65L12, 65L20, 65L70}

\maketitle

\section{Introduction}\label{sec1}
Central venous catheterization (CVC) is widely performed in intensive care, trauma, major surgery, parenteral nutrition, and hemodialysis, with over 7 million annual placements in the U.S. \cite{kehagias2023central}. It is typically performed at the internal jugular (IJ), subclavian, or femoral veins, with IJ preferred for its superficial location, consistent anatomy, and reduced risk. Ultrasound-guided CVC is now the clinical standard \cite{govindan2018peripherally}. The procedure involves ultrasound scanning to identify vessels, estimating insertion point, and advancing the needle. Major complications occur in 3\% of placements (i.e., $4.4$ pneumothorax, $16.2$ arterial puncture, $2.8$ arterial cannulation per 1000 cases) \cite{teja2024complication}. Despite ultrasound guidance, continuous simultaneous visualization of 
a vessel and needle is difficult due to narrow imaging planes and { vessel “roll-away” during advancement}. %, even for experts. UNCLEAR WHY THIS IS NEEDED. AS OPPOSED TO?
% Although ultrasound guidance reduces complications, maintaining continuous visualization of both the vessel and advancing needle in the narrow imaging plane remains a major challenge \cite{ikhsan2018assistive}. 
% Misalignment between the probe and needle can obscure the tip, increasing the risk of complications such as arterial puncture, hematoma, or pneumothorax. 
% Even experienced practitioners have to carefully balance probe positioning and needle advancement. { In addition, vessel “roll-away” during needle advancement further complicates access.} 
These challenges highlight the need for advanced technologies to reduce workload and improve safety.

Robot-assisted ultrasound (RUS) has emerged as a promising solution to enhance precision, minimize human error, and improve outcomes in diagnostic and interventional applications \cite{biswas2023recent, raina2021comprehensive, raina2024coaching}. Several handheld or semi-automated needle insertion devices have been developed to improve workspace, accuracy, and force resistance \cite{slocum2014methods, cheng2019hand, ikhsan2018assistive}. For example, CathBot \cite{cheng2019hand} enables clinicians to initiate insertion by pushing a handle, after which the device autonomously completes cannulation. While effective for peripheral vascular access on the forearm, its design has limited CVC utility  due to anatomical complexities of neck, chest, or groin sites. { Other robotic systems demonstrated teleoperated insertion, initiated from manually marked regions \cite{zevallos2021toward, zevallos2025automatic} or vessel boundaries \cite{chen2021ultrasound}.} { Efforts have recently shifted towards autonomy with advanced robotics, imaging, and sensor-guided needles \cite{koskinopoulou2023robotic}, reinforced by the first-in-human image-guided system in \cite{herlihy2025centrally}, showing 100\% success with no complications in 19 attempts.} Similarly, Hong \textit{et al.} \cite{hong2004ultrasound} developed a percutaneous insertion robot compensating for patient motion via real-time visual-servoing. Neubach \textit{et al.} \cite{neubach2009ultrasound} introduced closed-loop steering using ultrasound-estimated tip position with tissue stiffness modeling. Xu \textit{et al.} \cite{xu2010three} proposed slice-guidance interpolating preoperative scans to compensate liver motion. Chatelain \textit{et al.} \cite{chatelain20153d} used particle filtering to track flexible needles for real-time closed-loop control. While these approaches \cite{hong2004ultrasound, neubach2009ultrasound, xu2010three, chatelain20153d} advance motion compensation, they lack integrated intra-operative planning for full autonomy.

Zettinig \textit{et al.} \cite{zettinig20173d} annotated pre-interventional plans in CT/MRI and proposed a 3D visual-servoing framework with multi-modal registration to update needle targets under motion.  
Kojcev \textit{et al.} \cite{kojcev2016dual} developed a dual-arm ultrasound-guided system integrating force-controlled acquisition, registration, vision-based control, and tracking, though planning still relied on manual region selection.  
Bell \textit{et al.} \cite{bell2018photoacoustic} and Gubbi \textit{et al.} \cite{gubbi2021deep} demonstrated photoacoustic-based visual servoing.  
Cheng \textit{et al.} \cite{cheng2016venipuncture} used impedance sensing to detect vascular entry.  
Koskinopoulou \textit{et al.} \cite{koskinopoulou2023dual} combined ultrasound and bioimpedance in a dual-arm system for vessel localization and venipuncture detection.  
While these systems \cite{zettinig20173d, bell2018photoacoustic, gubbi2021deep, kojcev2016dual, cheng2016venipuncture} demonstrate end-to-end procedures, scanning was typically initialized from predefined poses or required experts to review intra-operative CT/MRI images.  

In contrast, our system achieves fully autonomous scan initialization using feedback from a low-cost, widely accessible RGB-D camera, eliminating the need for expert intervention or expensive intra-operative imaging. In addition,  many of the initial system validations were performed with simplified vascular access phantoms that do not capture the anatomical and physiological complexity of humans. %, thereby limiting their clinical applicability. NOT QUITE TRUE, AS THIS IS OFTEN THE FIRST REQUIRED STEP
In this work, we validate our initial pipeline on a high-fidelity phantom model, moving one step closer to in-human procedures.
    {  Key contributions of this work include:
    \begin{enumerate}
        \item First robotic pipeline that autonomously performs key CVC steps: intra-operative planning, scanning, anatomical localization, and needle placement
    
        \item A novel method to predict the optimal scanning region and trajectory using only a single depth image from an RGB-D camera.
    
        \item A fully automated robotic scanning and vascular reconstruction for localization of vein and artery, followed by supervised needle guidance.
    
        \item Feasibility demonstration of the system on a near-realistic CVC phantom for IJ vein access across 10 diverse clinical scenarios.
    \end{enumerate}}

\section{Materials and Methods}
% This section will give an overview of the workflow for the proposed autonomous robotic system for CLP procedure. 
% The subsection are organized as follows: 
\subsection{Experimental testbed} \label{sec:exp_testbed}
% \begin{figure}[h]
%     \centering
% 	%trim={L,B,R,T}
% 	\includegraphics[trim=0cm 6cm 4cm 0cm,clip,width=\linewidth]{figs/exp_setup_correct}
%     \caption{Overview of the experimental testbed showing the robotic system setup, a close-up view of the needle insertion mechanism, and the defined reference frames. The corresponding homogeneous transformation matrices, denoted as $\boldsymbol{T}_1^2$, represent the pose of frame 1 relative to frame 2.}
%     \label{fig:exp_setup}
% \end{figure}
    \begin{figure}[h]
    \centering
    	%trim={L,B,R,T}
        \includegraphics[trim=0cm 6cm 4cm 0cm,clip,width=\linewidth]{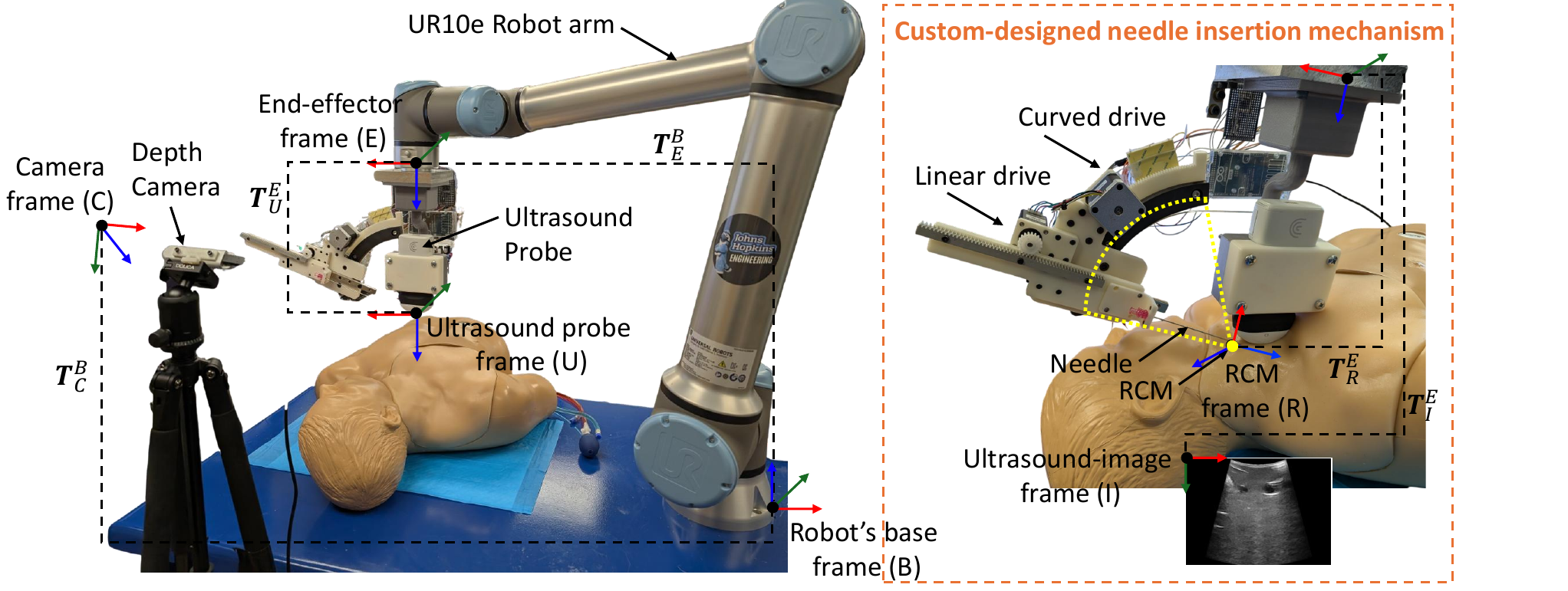}
        \caption{Overview of the experimental testbed showing the robotic system setup, a close-up view of the needle insertion mechanism, and the defined reference frames. The corresponding homogeneous transformation matrices, denoted as $\boldsymbol{T}_1^2$, represent the pose of frame 1 relative to frame 2.}
        \label{fig:exp_setup}
    \end{figure}
% The experimental testbed (Fig. \ref{fig:exp_setup}) includes: (1) a $6$-DoF UR10e robotic arm (Universal Robots, Denmark) with $10$ kg payload and inbuilt force-torque sensor, (2) a C3 HD3 curved ultrasound probe (Clarius, Canada) mounted via an extended attachment for real-time imaging. Although linear probes are preferred for CVC, we used a convex probe for its wider field of view and availability during prototyping. (3) An Ubuntu PC and a laptop, with NVIDIA RTX 4090 (16GB) and 1090 (12GB) for system control and deep learning inferences, respectively, (4) an Intel RealSense D415 RGB-D camera (California, USA) for spatial localization, (5) a custom 2-DoF needle insertion mechanism, adapted from the design in \cite{chen2021ultrasound}, providing linear and angular control, compatible with any robotic arm with adequate payload capacity, and equipped with an 18-gauge, 15 cm percutaneous thin-wall bevel-tipped needle (Cook Medical, USA). { Unlike manual procedures that rely on clinician skill to infer the needle tip, this design aligns the ultrasound plane and needle trajectory, ensuring continuous visualization throughout insertion.} (6) The CVC Blue Phantom (CAE, USA) for validation, offering highly realistic upper thorax and neck anatomy, including major vessels and anatomical landmarks commonly used in CVC. 
The experimental testbed (Fig. \ref{fig:exp_setup}) includes: (1) a $6$-DoF UR10e robotic arm (Universal Robots, Denmark) with $10$ kg payload and inbuilt force-torque sensor; (2) a C3 HD3 curved ultrasound probe (Clarius, Canada) mounted via extended attachment (although linear probes are preferred for CVC, we used a convex probe for its wider field of view and availability during prototyping); (3) an Ubuntu PC and a laptop with NVIDIA RTX 4090 (16GB) and 1090 (12GB) for system control and deep learning-based feedback; (4) a RealSense D415 RGB-D camera (Intel, USA) for localization; (5) a custom 2-DoF needle insertion mechanism adapted from \cite{chen2021ultrasound}, providing linear and angular control, compatible with any robotic arm with adequate payload, equipped with an 18-gauge, 15 cm bevel-tipped needle (Cook Medical, USA) {  to constantly align its trajectory with the ultrasound plane, ensuring continuous visualization (unlike manual procedures, which rely on clinician skill to infer the needle tip); and } (6) a CVC Blue Phantom (CAE, USA) for validation, offering realistic upper thorax and neck anatomy, including major vessels and landmarks for CVC.
Note that the communication across system components was established via ROS2 (Robot Operating System).

Let $\boldsymbol{T}_1^2$ denote the homogeneous transformation matrix that expresses the pose of frame 1 relative to frame 2. { The frames used in the system are: robot's base (B), end-effector (E), camera (C), ultrasound probe (U),  remote center-of-motion (R), and ultrasound image (I) (see Fig. 1)}. 
% The world coordinate system is defined at the robot base. 
% $\boldsymbol{T}_B^E$ defines the transformation of robot's end effector with respect to the robot's base and was estimated using the Denavit–Hartenberg (DH) parameters provided by the manufacturer. $\boldsymbol{T}_B^C$ defined the transformation of 3D camera with respect to the robot's base and was estimated using the hand-to-eye calibration method. This involved capturing multiple paired observations of the robot's end-effector poses and the corresponding camera frames for the same 3D point in the environment, followed by solving an optimization problem to determine the optimal transformation matrix that aligned the camera coordinate frame with the robot's base frame. $\boldsymbol{T}_E^P$ and $\boldsymbol{T}_E^{C}$ defined the transformation of ultrasound probe tip and remote center-of-motion (RCM), respectively, with respect to the robot's end-effector. These transformations were rigidly defined based on the CAD model. $\boldsymbol{T}_E^U$ defined the transformation of ultrasound image with respect to the robot's end-effector and is estimated using the cross-wire phantom. This involved acquiring multiple ultrasound images of the cross-wire phantom and the corresponding end-effector poses. Since the location of cross-wire was fixed relative to the base of robot, thus equation $(\boldsymbol{T}_B^{cross})_1 = (\boldsymbol{T}_B^{cross})_2$ can be rearranged as AX=XB problem to find the $\boldsymbol{T}_E^U$. 
The 3D position in world coordinate system ($p_B$) of a given pixel in the 2D ultrasound image ($p_I$) is calculated using:
% \begin{equation} \label{eq:pixel_to_3d}
%     \boldsymbol{p}_B = \boldsymbol{T}_B^E . \boldsymbol{T}_E^U . \boldsymbol{T}_U^I . \boldsymbol{p}_I
% \end{equation}
\begin{equation} \label{eq:pixel_to_3d}
    \boldsymbol{p}_B = \boldsymbol{T}_E^B . \boldsymbol{T}_U^E . \boldsymbol{T}_I^U . \boldsymbol{p}_I
\end{equation}
where $\boldsymbol{T}_I^U$ is the transformation from the ultrasound probe frame to the pixel values in the image frame and is given by
\begin{equation}
    \boldsymbol{T}_I^U =
    \begin{bmatrix}
        S_x & 0 & 0 & 0 \\
        0 & S_y & 0 & 0 \\
        0 & 0 & 1 & 0 \\
        0 & 0 & 0 & 1
    \end{bmatrix}
    \begin{bmatrix}
        I_x \\ I_y \\ 0 \\ 1
    \end{bmatrix}
\end{equation}
where $S_x = S_y = 0.15436~mm$ is the pixel spacing (scaling factor) provided by the Clarius Mobile Health software development kit and $(I_x, I_y)$ is the given pixel value in $x-$ and $y-$direction of the image frame. 
% The formulation in eq. \eqref{eq:pixel_to_3d} was used for 3D reconstruction of the anatomical structures after the robotic scanning. 

\subsection{Pipeline} \label{sec:pipeline}
\begin{figure}[h]
    \centering
	%trim={L,B,R,T}
	% \includegraphics[trim=0cm 16.3cm 0cm 0cm,clip,width=\linewidth]{figs/pipeline}
	\includegraphics[trim=0cm 11.3cm 0cm 0cm,clip,width=\linewidth]{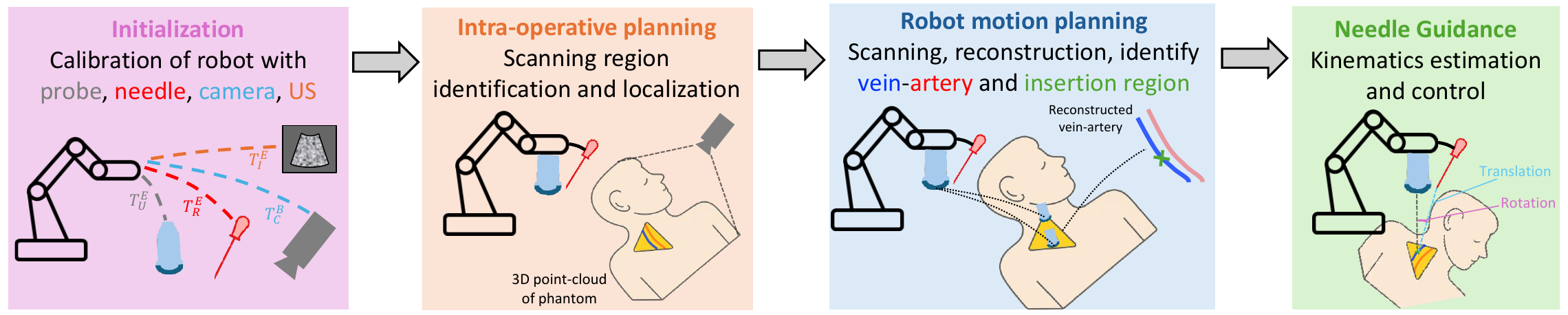}
    \caption{Our pipeline has four phases: the initialization phase , intra-operative planning, robot motion planning, and the needle guidance}
    \label{fig:workflow}
\end{figure}

Our pipeline is shown in Fig. \ref{fig:workflow}. During initialization, system calibration was performed by estimating $\boldsymbol{T}_C^B$ via hand-eye calibration and $\boldsymbol{T}_U^E$ using the classical $AX = XB$ method \cite{kim2013ultrasound}, with a cross-wire phantom in a water bath. The region encompassing IJ vein and carotid artery was then autonomously identified from depth images, followed by robotic ultrasound scanning, vessel segmentation, 3D reconstruction, insertion point selection, and closed-loop needle guidance for puncture.
% Our pipeline is shown in Fig. \ref{fig:workflow}. During the initialization phase, the system was calibrated by estimating the transformations $\boldsymbol{T}_C^B$  using hand-eye calibration method and $\boldsymbol{T}_E^U$ using the classical $AX = XB$ formulation \cite{kim2013ultrasound}. The latter was performed by using a cross-wire phantom submerged in a water bath. Then, a scanning region encompassing the IJ vein and carotid artery was autonomously identified using depth images, followed by robotic ultrasound scanning, vessel segmentation, 3D reconstruction, insertion point selection, and closed-loop needle guidance for puncturing the vein.
% The process begins with inputting a patient's point cloud into a deep regression network, which will predict clinically relevant anatomical landmarks. 
\subsection{Intra-operative planning} \label{sec:scanning_path}
% Fig. \ref{fig:scan_region_flow} shows the flowchart for scanning region identification and localization. 
\subsubsection{Overview}
Clinicians often define a triangular region using the suprasternal notch (SN), sternocleidomastoid (SCM) midpoint, and lateral clavicle (LC), known as Sedillot’s triangle, to guide IJ vein access \cite{kosnik2024anatomical}. Inspired by this, we used SN, LC, and anterior hyoid bone tip (HT) to define a similar region, with HT chosen for its prominence and absence of SCM in the phantom. This region ensured presence of the IJ vein. These Landmarks were identified using a deep learning model, which were then used to plan the scanning path, as described below.
% Clinicians often use a triangular region defined by the supersternal notch (SN), sternocleidomastoid (SCM) midpoint, and lateral clavicle (LC), known as Sedillot’s triangle, to guide IJ vein access \cite{kosnik2024anatomical}. Following this clinical protocol, we defined a similar scanning region using the SN, LC, and anterior hyoid bone tip (HT), the latter chosen for its anatomical prominence and absence of SCM in the CLP phantom. The region formed by SN, LC and HT still ensured the presence of the IJ vein. In order to identify these landmarks, we used a deep learning model, Dense U-Net, as explained below.
% An HSV-based thresholding technique is then applied to the color image to segment the region of interest (RoI), specifically the patient's head and torso. The segmented RoI is used to extract corresponding depth data, which is subsequently transformed into a 3D point cloud using the camera's intrinsic parameters and deprojection model. Depth filtering is also applied to eliminate extraneous points, ensuring a refined point cloud representation of the scene. 
% The processed point cloud is then passed on to the Dense U-Net model to identify anatomical landmarks and generate the scanning path, as explained in the subsequent sections. 
% \begin{figure}[h]
%     \centering
% 	%trim={L,B,R,T}
% 	\includegraphics[trim=0cm 13.5cm 15cm 0cm,clip,width=0.8\linewidth]{figs/scanning_region_flow}
%     \caption{Flowchart for estimating scanning region and path}
%     \label{fig:scan_region_flow}
% \end{figure}
% \subsection{Scanning path estimation using anatomical landmarks}

\subsubsection{Anatomical landmark prediction network}
% The anatomical landmark prediction network required a depth image of neck and torso region of the body to predict the landmarks. First, a synchronized depth and color images of the environment was captured by the depth camera. An HSV-based thresholding then segmented the region-of-interest i.e. phantom's head and torso, followed by depth filtering for a refined 3D point cloud. The segmented RoI point cloud was provided as input to the network, as shown in Fig. \ref{fig:scan_region_phantom}(a).  
% % The input of the network was the depth image that is computed from a segmented RoI point cloud, as shown in Fig. \ref{fig:scan_region_phantom}(a). 
% % During pre-operative planning, 

\begin{figure}[h]
    \centering
	%trim={L,B,R,T}
	\includegraphics[trim=0cm 12cm 0cm 0cm,clip,width=\linewidth]{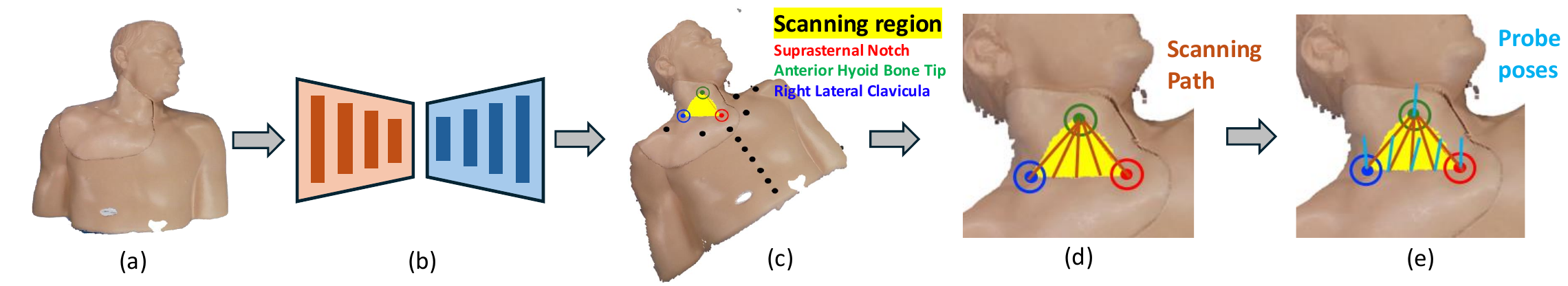}
    \caption{(a) Depth image of the phantom's neck and torso. (b) Landmark prediction network for anatomical localization. (c) Scanning region (yellow patch) on the neck is defined using three predicted anatomical landmarks. (d) Planned scanning paths are shown as maroon lines. (e) Probe poses along the scanning paths, with surface normals at each pose illustrated by light-blue lines.}
    \label{fig:scan_region_phantom}
\end{figure}

% Following the approach in \cite{teixeira2023automated}, we trained a Dense-UNet \cite{li2018h} to predict the location of internal landmarks from computed depth images. The network architecture consisted of an encoder and decoder, each with four dense blocks of four convolutional layers (Fig. \ref{fig:scan_region_phantom}(b)). Batch normalization and ReLU activation were applied after each convolutional layer, except for the final layer. 
The anatomical landmark prediction network required a depth image of the neck–torso region. Synchronized depth and color images were first captured, followed by HSV thresholding to segment the head and torso of the phantom, then depth filtering for a refined 3D point cloud. This segmented point-cloud was  the network's input (Fig. \ref{fig:scan_region_phantom}(a)). Following \cite{teixeira2023automated}, we trained a Dense-UNet \cite{li2018h} to predict internal landmarks from depth images. The encoder–decoder architecture had four dense blocks of four convolutional layers (Fig. \ref{fig:scan_region_phantom}(b)), with batch normalization and ReLU after each convolution except the final layer.
% The dataset of 4,912 depth images was used to train the Dense-UNet model, consisting of real human subjects with diverse body habitus, sex, and skin tone profiles. Each image was paired with manually annotated 3D anatomical landmarks by trained technicians using co-registered RGB-D data and verified against physical anthropometric measurements to ensure accuracy. The network was trained for 300 epochs, minimizing the Adaloss function using the using the Adam optimizer (learning rate = 0.001, batch size = 32).
The dataset comprised 4,912 depth images of real humans with diverse habitus, sex, and skin tone. Landmarks were manually annotated using co-registered RGB-D data and verified against anthropometric measurements. The model was trained for 300 epochs with Adam optimizer (lr = 0.001, batch = 32) minimizing Adaloss, predicting 33 landmarks, including three critical to CVC, with additional landmarks  to improve accuracy.
% for regularization The model was trained on a dataset of 4,912 images from real humans using the Adam optimizer (learning rate = 0.001, batch size = 32) for 300 epochs, minimizing the Adaloss function \cite{teixeira2019adaloss}. 
% The network predicted 33 landmarks, using additional ones for regularization to enhance accuracy of the 3 landmarks critical to CVC.
% which includes the Suprasternal notch, Anterior hyoid bone tip, Aortic arch center, Brachiocephalic artery bifurcation, C3-C7 vertebrae, Right/Left rib insertions, Carotid bifurcations, Lateral clavicles, Lung tops, Medial clavicles, Posterior hyoid bone tip, Subclavian and vertebralis branches, Vertebral artery C5, and T1-T12 vertebrae. 

\subsubsection{Scanning region and path estimation}
% Clinicians often use a triangular region defined by the supersternal notch (SN), sternocleidomastoid midpoint, and lateral clavicle (LC), known as Sedillot’s triangle, to guide IJ vein access \cite{kosnik2024anatomical}. Following this clinical protocol, we defined a similar scanning region using the SN, LC, and anterior hyoid bone tip (HT), the latter chosen for its anatomical prominence. 
The scanning region was defined by a triangular region formed by 3 predicted landmarks (SN, LC, and HT). This region was projected onto the phantom surface using the Open3D Python library, as shown in Fig. \ref{fig:scan_region_phantom}(c). To scan within this region, we defined straight-line paths from the triangle's base (SN–LC) to its apex (HT), aligning with the IJ vein's orientation. Initial paths connected HT to LC and HT to SN. Additional scanning paths were created by dividing the SN–LC line into equal intervals and connecting lines from HT to these points, yielding a dense scan pattern for accurate vessel detection (Fig. \ref{fig:scan_region_phantom}(d)).  
% These paths were also projected onto the phantom using Open3D (Fig. \ref{fig:scan_region_phantom}(d)). 
% Scanning began from the midpoint-to-HT path, with others used only if needed for reliable vessel reconstruction. 
Scanning was initiated along the midpoint-to-HT path because this path aligned with the expected anatomical course of the IJ vein, offering a clinically representative initial view. This path also provided a reliable starting point across different patient positioning due to its geometric symmetry.
\subsection{Robot motion planning}
\subsubsection{Overview}
\begin{figure}[h]
    \centering
	%trim={L,B,R,T}
	\includegraphics[trim=0cm 11cm 12cm 0cm,clip,width=0.8\linewidth]{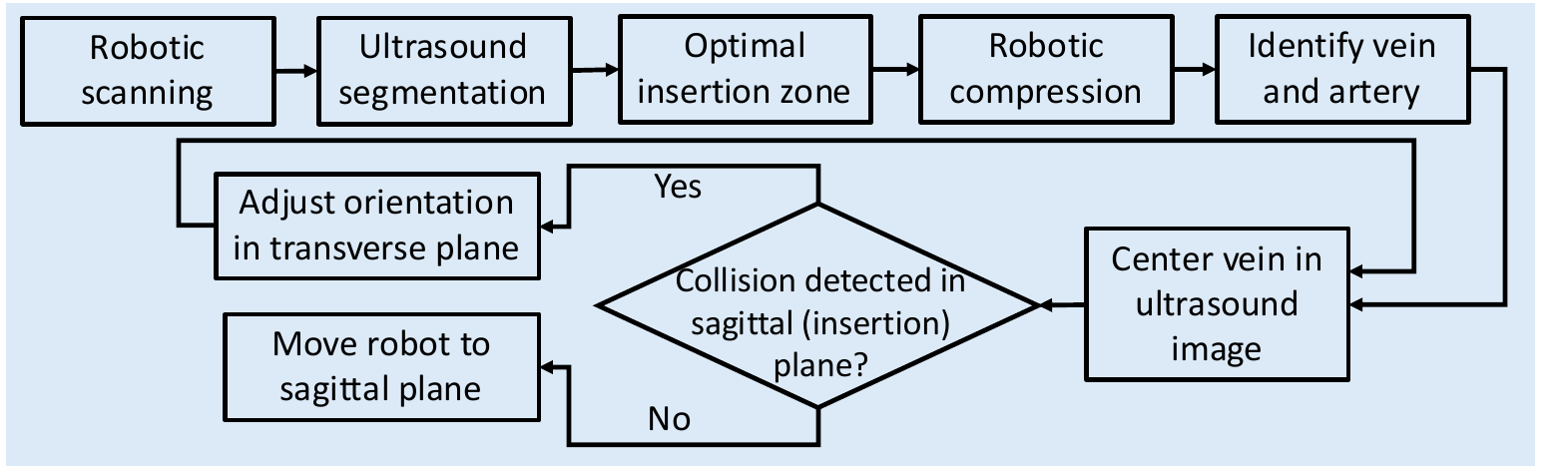}
    \caption{Robot motion planning worlflow to identify optimal insertion point, localize vein-artery, and move to sagittal (insertion) plane while avoiding collision.}
    \label{fig:robot_motion_planning}
\end{figure}

Once the scanning region was identified, the pipeline generated a robot trajectory for scanning and localizing the vein-artery, as illustrated in Fig. \ref{fig:robot_motion_planning}. After robotic scanning, the vein was localized and centered in the ultrasound image to ensure consistent spatial referencing. Then the robot was re-oriented for sagittal plane insertion, which offered full needle shaft visibility. A collision check was performed using the CAD model and phantom point cloud. If a collision was detected in the sagittal orientation, the robot iteratively adjusted its pivot angle in transverse view about the point of contact by $5^\circ$ until a collision-free pose was found. Upon confirming clearance, the robot transitioned to the sagittal plane for needle insertion.

% The motion planning algorithm first generates a trajectory of robot poses on the phantom surface. The robot will then scan along this trajectory and acquire US images, which will be processed to reconstruct the vascular structures and identify the optimal insertion point while also localizing the vein-artery and moving to the insertion (sagittal) plane. The flowchart of an algorithm is given in Fig. \ref{fig:robot_motion_planning}.  
\subsubsection{Scanning trajectory} \label{sec:scan_path_plan}
The scanning path generated on the phantom in Section~\ref{sec:scanning_path} defined the robot trajectory as:
\begin{equation}
    \boldsymbol{\mathcal{T}} = [\boldsymbol{P}_i, \cdots, \boldsymbol{P}_n]
\end{equation}
Each pose transform $\boldsymbol{P}_*$ was defined as $[\overrightarrow{\boldsymbol{x}}, \overrightarrow{\boldsymbol{y}}, \overrightarrow{\boldsymbol{z}}, \overrightarrow{\boldsymbol{t}}]$, where $\overrightarrow{\boldsymbol{t}}$ is the translation (or position) vector and $\overrightarrow{\boldsymbol{x}}, \overrightarrow{\boldsymbol{y}}, \overrightarrow{\boldsymbol{z}}$ are unit vectors for orientation. The start and end of the scanning path define $\boldsymbol{P}_0$ and $\boldsymbol{P}_n$, respectively. The normal $\overrightarrow{\boldsymbol{z}}$ of phantom was computed using Open3D and aligned with the end-effector's normal $\overrightarrow{\boldsymbol{z}_e}$ using:
\begin{equation}
\overrightarrow{\boldsymbol{z}} = 
\begin{cases} 
\overrightarrow{\boldsymbol{z}}, & \text{if } \overrightarrow{\boldsymbol{z}} \cdot \overrightarrow{\boldsymbol{z_e}} > 0 \\ 
-\overrightarrow{\boldsymbol{z}}, & \text{otherwise} 
\end{cases}
\end{equation}
The x-axis $\overrightarrow{\boldsymbol{x}}$ was aligned with the phantom's major axis, and $\overrightarrow{\boldsymbol{y}}$ was computed as:
\begin{equation}
    \overrightarrow{\boldsymbol{y}} = \frac{\overrightarrow{\boldsymbol{z}} \times \overrightarrow{\boldsymbol{x}}}{||\overrightarrow{\boldsymbol{z}} \times \overrightarrow{\boldsymbol{x}}||}
\end{equation}
Intermediate poses were interpolated between $\boldsymbol{P}_0$ and $\boldsymbol{P}_n$ using the UR driver. ROS2 and URScript executed the trajectory in hybrid force-position mode with $5$ N force along the $z$-axis of the probe{ , chosen as the minimum force required to ensure sufficient acoustic contact while minimizing tissue deformation.} During scanning, a rosbag recorded poses and images pairs $(\boldsymbol{P}_i, \boldsymbol{I}_i)$ for each frame,  described as follows to enable accurate vessel reconstruction:
\begin{equation} \label{eq:rosbag_data}
    \boldsymbol{D} = \{(\boldsymbol{P}_i, \boldsymbol{I}_i), \cdots, (\boldsymbol{P}_n, \boldsymbol{I}_n)\} 
\end{equation}

\subsubsection{Identifying optimal insertion plane}
\begin{figure}[h]
    \centering
	\includegraphics[trim=0cm 9cm 18cm 0cm,clip,width=0.7\linewidth]{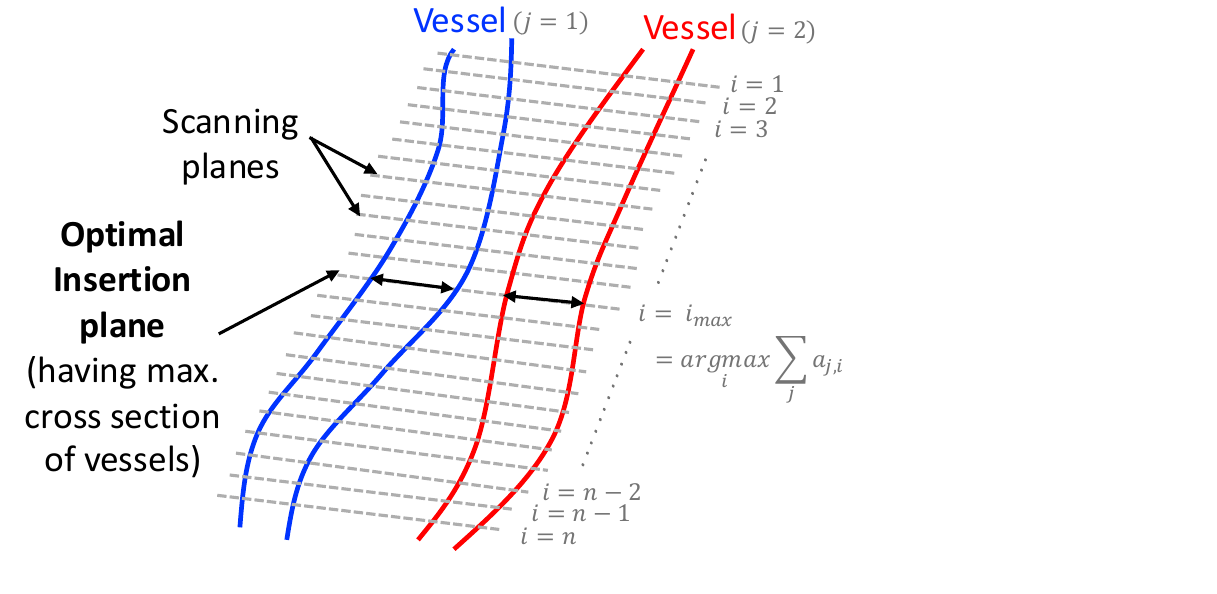}
    \caption{Criteria for selecting optimal needle insertion plane. $a_{j,i}$ is the area of the $j^{th}$ vessel in image $I_i$ from dataset $\boldsymbol{D}$ (Eq.~\eqref{eq:rosbag_data}), with $j$ limited to 2 to exclude noise, as only two vascular structures are expected to be visible.}
    \label{fig:vein_artery_classi}
\end{figure}

The acquired 2D US frames were segmented to reconstruct vessels and plan needle insertion. A deep learning model from prior work \cite{al2025robotic} was used to extract vessel center-points and areas from each frame. Ideally, insertion should occur where both vessels (vein and artery) are clearly visible for confident and precise targeting. To determine the insertion point, the robot insertion pose ($\boldsymbol{\mathcal{P}}$) was selected from $\boldsymbol{P}_i, \cdots, \boldsymbol{P}_n$ as that corresponding to the maximum combined area of the two vessels. This selection criterion is illustrated in Fig. \ref{fig:vein_artery_classi}.
% Mathematically, it can be represented as:
% \begin{equation}
%     \boldsymbol{\mathcal{P}} = \boldsymbol{P}_{(i=i_{max})}, \quad
%     i_{max} = \arg\max_{i} \sum_{j=1}^{m} a_{j,i}
% \end{equation}
% Here, $a_{j,i}$ is the area of the $j^{th}$ vessel in image $I_i$ from dataset $\boldsymbol{D}$ (Eq.~\eqref{eq:rosbag_data}), with $j$ limited to 2 to exclude noise, as only two vascular structures are expected to be visible. 
The robot was finally moved to pose $\boldsymbol{\mathcal{P}}$, and the corresponding ultrasound frame was re-segmented.
\subsubsection{Vein-artery localization}
Veins are difficult to distinguish from arteries, as both appear morphologically similar in ultrasound images. While clinicians typically use Doppler ultrasound to differentiate vessels based on blood flow, this method requires manual tuning, favorable imaging conditions, and expertise, making it less practical in portable or low-cost devices where Doppler functionality or detectable flow is absent. To address these limitations, our system employs a compression-based method that exploits differences in vascular compliance, whereby veins exhibit greater deformation than arteries under low external pressure.

The robot applied controlled compression by gradually increasing the contact force from $5$ N to $10$ N, then decreasing back to $5$ N, while recording an ultrasound video. {  Note that this compression was applied only briefly for classification. The maximum  applied force of 10 N is considered clinically safe and does not induce patient discomfort.} Segmented vessel areas and centroids were extracted from the recorded sequence. Then, KMeans clustering was used, which grouped vessel centroids into two clusters based on their spatial proximity. Area changes across the compression cycle were computed for each cluster, with the one showing greater deformation identified as the vein, and the other as the artery. The centroid of the vein cluster was stored as a reference. Later, at the insertion pose, the robot captured a new ultrasound frame, segmented the vessels, and identified the vein by finding the centroid closest to the saved reference, with the remaining vessel labeled as the artery.
\subsection{Needle guidance} \label{sec:needle_guidance}
{ The robotic needle mechanism has two degrees of freedom: (i) rotation ($\theta$) to align the needle with the desired insertion angle, and (ii) translation ($d$) to advance the needle to the target depth. These desired angle and depth of needle are computed from the kinematic relationship between the IJ vein center and the RCM of the needle mechanism (see Fig.~\ref{fig:exp_setup}).

    Let $p^{VC} = [x^{VC}, y^{VC}, z^{VC}]$ and $p^{RCM} = [x^{RCM}, y^{RCM}, z^{RCM}]$ denote the 3D coordinates of the vein center and RCM, respectively, in the robot base frame. The vein center was first identified from the segmented region in the transverse ultrasound plane and then tracked in the sagittal plane. The desired insertion angle $\theta$ is defined as the angle between the line connecting the RCM to the vein center and the vertical $z$-axis of the robot's end-effector frame, with an offset to account for the mechanism design:
    
    \begin{equation}
    \theta = \tan^{-1}\left(\frac{x^{VC} - x^{RCM}}{z^{VC} - z^{RCM}}\right) - \theta_{\text{offset}},
    \end{equation}
    where $\theta_{\text{offset}} = 18.21^{\circ}$ is the initial tilt of the needle guide relative to the robot $z$-axis. The insertion depth is given by the Euclidean distance between the vein center and the RCM:
    
    \begin{equation}
    d = | p^{VC} - p^{RCM}|.
    \end{equation}}
    
    To minimize targeting error, a closed-loop controller was implemented. The controller adjusted the  translational motion of the needle to minimize the perpendicular distance between the needle tip and pre-defined vein centerline during advancement. Since the needle orientation was fixed prior to insertion, this control ensured safe puncture and in-vein positioning. The proportional controller used a gain of 0.5. At present, the vein centerline was marked manually once in the sagittal ultrasound view, and the needle tip location was manually identified after each advancement step. The needle was advanced until the tip reached within 1~mm of the centerline.
\section{Experiments and Results}
\subsection{Overview}
\begin{figure}[h]
    \centering
	%trim={L,B,R,T}
	\includegraphics[trim=0cm 13cm 4.5cm 0cm,clip,width=\linewidth]{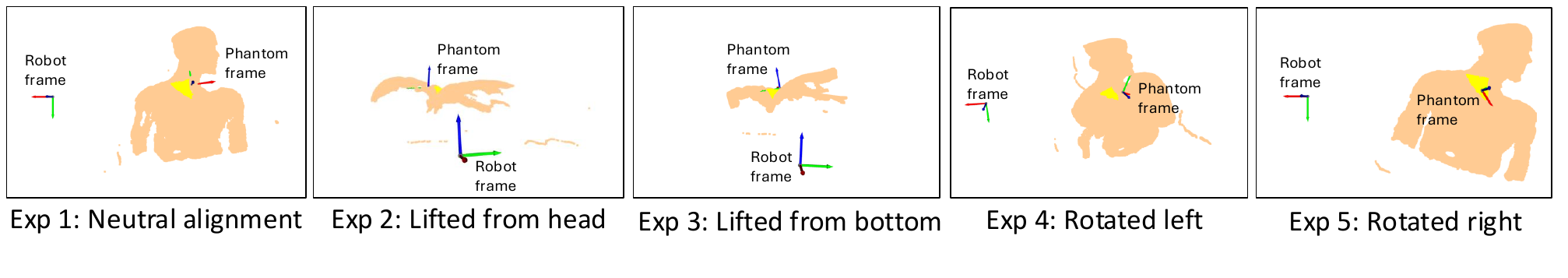}
    \caption{Illustrating experiments 1–5, each with a different phantom orientation relative to the robot base to assess system's robustness to patient positioning in different clinical scenarios.}
    % \caption{orientations of phantom frame with respect to robot base frame for experiments 0-4.}
    \label{fig:expset1_frames}
\end{figure}

To comprehensively evaluate the robustness and clinical relevance of our system, we conducted 10 experiments under varied simulated clinical scenarios. These included: (i) five experiments (Exp. 1-5) with different phantom orientations representing common patient positions during CVC (e.g., Trendelenburg, head turns, torso tilt); and (ii) five additional experiments (Exp. 6-10) introducing random perturbations by adding Gaussian noise ($\mu=0$, $\sigma=0.01$) to the scanning path to simulate real-world uncertainties such as patient motion or breathing. 
Fig. \ref{fig:expset1_frames} illustrates the five baseline orientations, while Table \ref{tab:experiment_details} summarizes the corresponding phantom-to-robot frame transformations and noise-induced deviations. 

\begin{table}[h]
    \caption{Exp. 1-5 corresponds to scenarios with different orientations (in $deg.$) of phantom with respect to (w.r.t) robot base and Exp. 6-10 corresponds to scenarios with different amounts of deviations (in $mm$) added to the start and end point of the scanning path.}
        \centering
    \begin{tabular}{ccccccccc}
        \toprule
         \multirow{2}{*}{Exp.} & \multicolumn{3}{c}{Rotation of phantom w.r.t robot} & \multirow{2}{*}{Exp.} & \multicolumn{2}{c}{Start Deviation} & \multicolumn{2}{c}{End Deviation} \\
        \cmidrule(lr){2-4} \cmidrule(lr){6-7} \cmidrule(lr){8-9}
        & Roll & Pitch & Yaw &  & $\Delta x$ & $\Delta y$ & $\Delta x$ & $\Delta y$ \\
        \midrule
        1 & -3.220 & -1.711 & -173.129 & 6 & +3.295 & -4.298 & -8.128 & +7.024 \\
        2 & 6.011 & -0.376 & -177.914 & 7 & +5.385 & -17.523 & +23.160 & -5.776 \\
        3 & 13.003 & -0.065 & -174.284 & 8 & +3.024 & -5.065 & -9.509 & +1.549 \\
        4 & 4.036 & -1.779 & 122.572 & 9 & -1.270 & +6.466 & -1.229 & -16.543 \\
        5 & 3.016 & 3.377 & 155.136 & 10 & -10.250 & +6.343 & -2.225 & -15.113 \\
        \bottomrule
    \end{tabular}
    \label{tab:experiment_details}
\end{table}

\subsection{Landmark prediction}
% With our testing set of 427 depth images, the mean Euclidean error for  33 total landmarks detected was $13.59$ $mm$ ± $10.49$ $mm$ with a $95^{th}$ percentile of $34.26$ $mm$. For three landmarks relevant to CVC, namely, the SN, RLC, and HT, a mean localization error of $13.88$ $mm$ ± $11.55$ is observed, with a $95^{\text{th}}$ percentile error of $35.35$ $mm$. Table \ref{tab:landmark_error} reports detailed results, highlighting the localization accuracy of critical anatomical landmarks. Although the maximum error reached $34.17$ $mm$, this did not significantly impair downstream vessel localization, as the triangular scanning region considered large enough to encompass the IJ vein across all test cases. However, reducing this error would further improve procedural accuracy, especially in anatomically complex patients.
With 427 test depth images, the mean Euclidean error for 33 landmarks was $13.59$ $mm$ $\pm$ $10.49$ $mm$ (95th percentile: $34.26$ $mm$). For the three key CVC landmarks, SN, RLC, and HT, the mean error was $13.88$ $mm$ ± $11.55$ (95th percentile: $35.35$ $mm$). Table~\ref{tab:landmark_error} shows detailed results. Although the maximum error reached $34.17$  $mm$, this error did not hinder downstream vessel localization, as the triangular scanning region was large enough to encompass the IJ vein. 
% Nonetheless, reducing this error could improve accuracy in anatomically complex patients.
\begin{table}[h]
    \centering
    % \caption{Mean error, standard deviation (S.D.), and 95th percentile error for anatomical landmarks of CLP.}
    \caption{Mean, standard deviation (S.D.), and 95th percentile error for CVC landmarks.}
    \begin{tabular}{cccc}
    \toprule
    % \textbf{Landmark} & \textbf{Mean error (mm)} & \textbf{S.D. (mm)} & \textbf{95\% Percentile (mm)} \\
    % \midrule
    % Suprasternal notch & 13.03 & 11.21 & 36.96 \\
    % Right lateral clavicula & 12.77 & 10.52 & 32.09 \\
    % Anterior hyoid bone tip & 15.84 & 12.91 & 37.01 \\
    % \midrule
    % \textbf{Average} & \textbf{13.88} & \textbf{11.55} & \textbf{35.35} \\
    \textbf{Landmark} & \textbf{Mean $\pm$ S.D. (95\textsuperscript{th} Percentile) [$mm$]} \\
    \midrule
    SN & 13.03 $\pm$ 11.21 (36.96) \\
    RLC & 12.77 $\pm$ 10.52 (32.09) \\
    HT & 15.84 $\pm$ 12.91 (37.01) \\
    \midrule
    \textbf{Average} & \textbf{13.88 $\pm$  11.55 (35.35)} \\
    \bottomrule
    \end{tabular}
    \label{tab:landmark_error}
\end{table}
\subsection{Vein-artery centerline reconstruction}
\begin{figure}[!h]
    \centering
	%trim={L,B,R,T}
	\includegraphics[trim=0cm 2cm 0cm 0cm,clip,width=\linewidth]{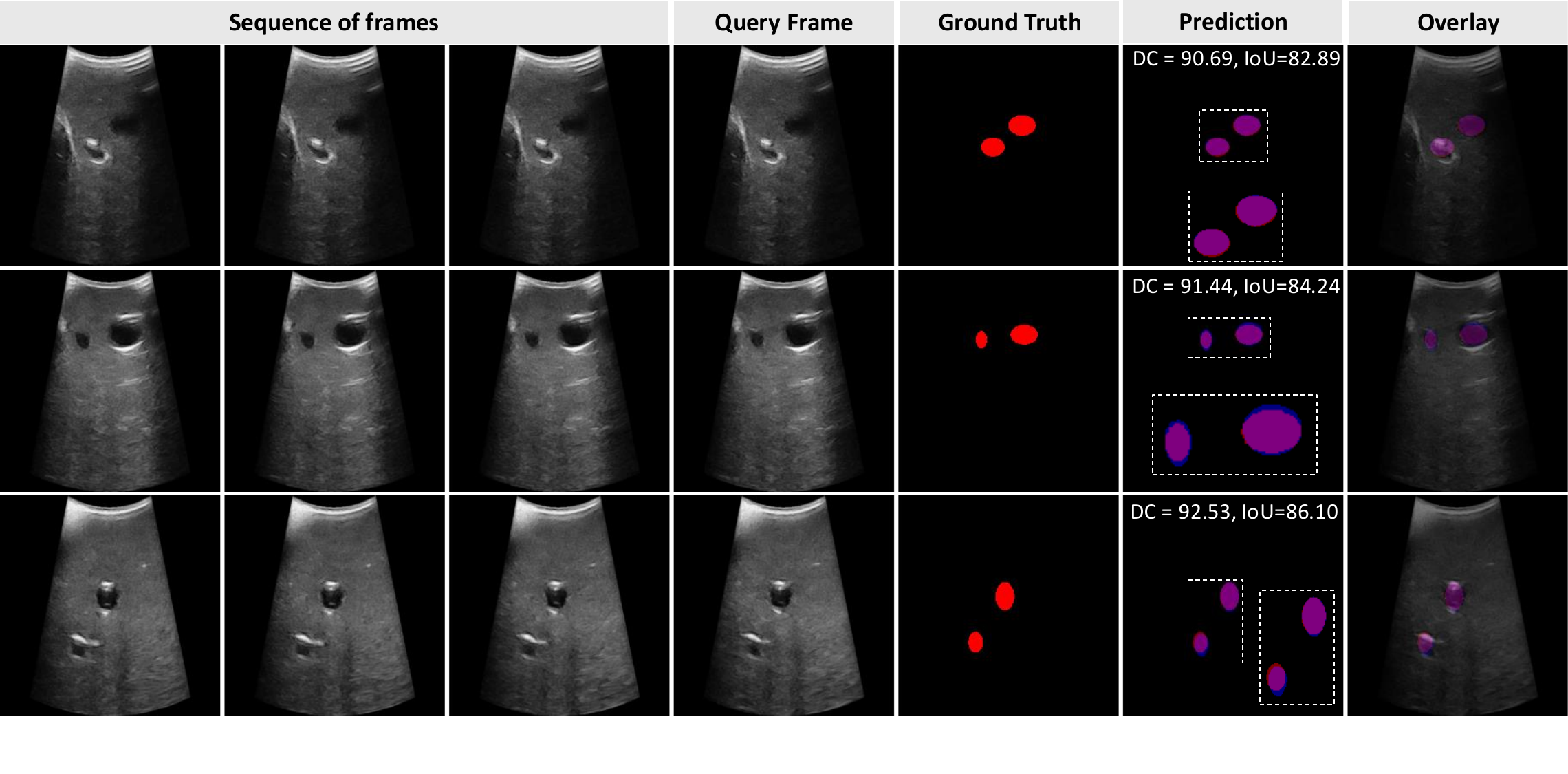}
    \captionsetup{labelfont={color=blue}, textfont={color=blue}}
    \caption{Comparison of segmentation network predictions with ground truth on the test set. Predicted regions are shown in blue, ground truth in red, and their overlap in magenta. White boxes highlight zoomed-in views of the overlapping regions.}
    \label{fig:us_seg}
    \end{figure}

\begin{figure}[!h]    
        \centering
    	%trim={L,B,R,T}
    \includegraphics[trim=0cm 0cm 11cm 0cm,clip,width=0.95\linewidth]{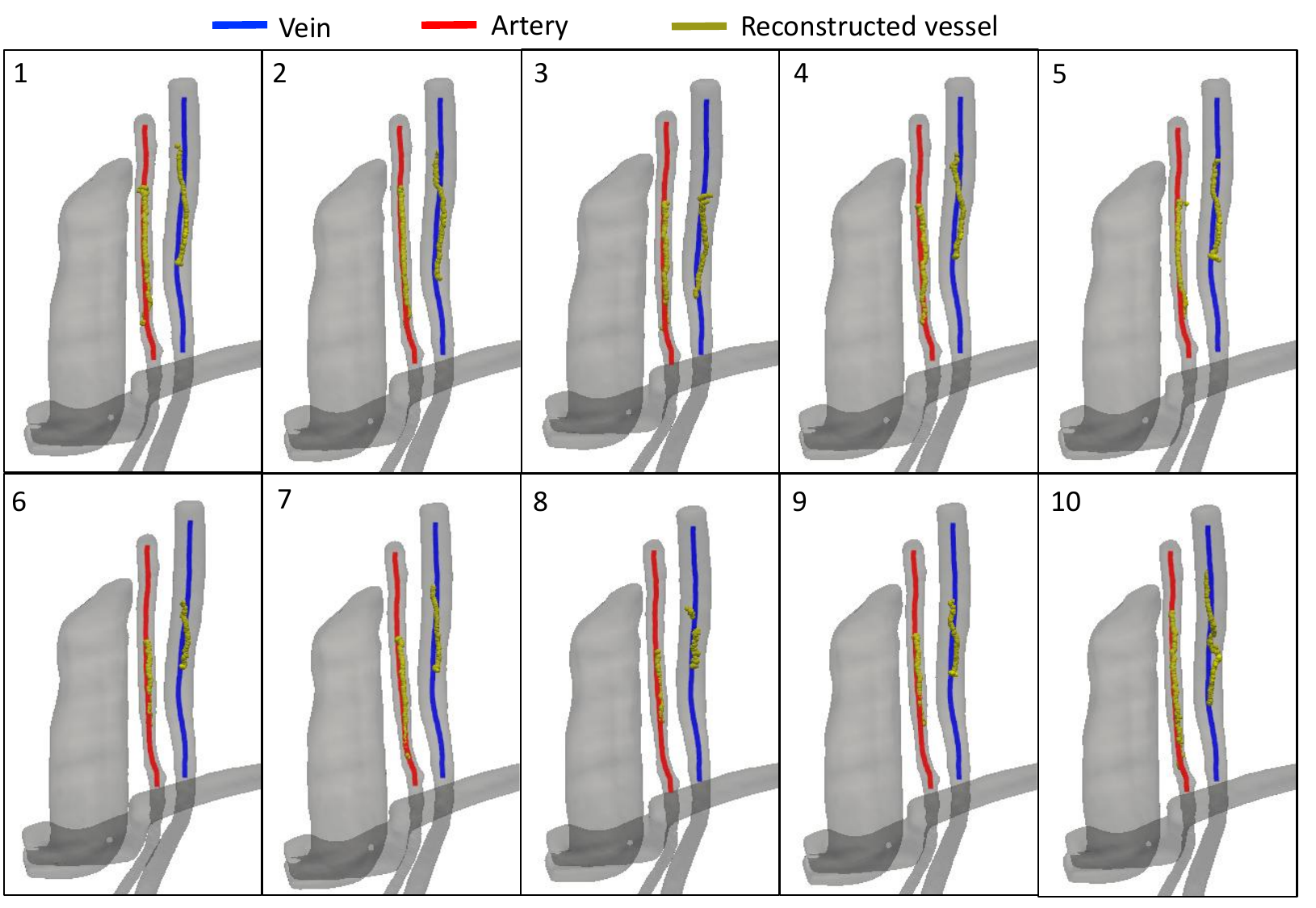}
        \captionsetup{labelfont={color=blue}, textfont={color=blue}}
        \caption{Comparison of 3D reconstruction of centerline of vein and artery (both marked with yellow color) with their ground truths (blue color for vein and red color for artery) for the 10 experiments}
        \label{fig:recon_results}
    \end{figure}

% \subsection{Experiments}
% Fig. \ref{fig:recon_results} shows examples of reconstructed centerlines of veins  and arteries (magenta) overlaid on corredponding ground truth centerlines for the vein (blue) and artery (red) %. 
{ Fig. \ref{fig:us_seg} shows representative segmentation results of veins and arteries from ultrasound frames. The reconstructed vein and artery centerlines were then obtained by transforming these segmentation outputs in the robot frame using Eq. \eqref{eq:pixel_to_3d} and compared with the ground truth centerlines extracted from CT scans of the phantom.} The CT data was processed using the Vascular Modeling Toolkit in 3D Slicer to extract the centerlines discretized at $1$ $mm$ intervals to serve as the ground truth. The reconstructed and ground truth centerlines were aligned using the ICP algorithm (Fig. \ref{fig:recon_results}). To assess the accuracy of reconstructed centerlines, we computed their $k-$nearest neighbors distance with the ground truth center-point, as reported in Table \ref{tab:compare_3d_va}.
% reports  computed distances per experiment, providing a quantitative assessment of the alignment between the reconstructed vessel and the actual anatomical data. 
\begin{table}[t]
    \centering
    \caption{Reconstruction accuracy reported as mean, maximum, and standard deviation (S.D.) errors (in $mm$) between reconstructed and ground-truth vein and artery center-lines.}
    \begin{tabular}{ccccccc}
    \toprule
        \multirow{2}{*}{Exp.} & \multicolumn{3}{c}{Vein} & \multicolumn{3}{c}{Artery} \\
       \cmidrule(l){2-4} \cmidrule(l){5-7}
        ~ & Mean & Max. & S.D. & Mean & Max. & S.D. \\
        \midrule
        1 & 3.28 & 5.06 & 0.98 & 1.21 & 2.24 & 0.43 \\
        2 & 3.13 & 5.05 & 0.96 & 1.05 & 2.49 & 0.47 \\
        3 & 2.74 & 4.56 & 0.78 & 1.08 & 2.49 & 0.39 \\
        4 & 3.32 & 4.93 & 0.96 & 1.20 & 2.32 & 0.44 \\
        5 & 3.28 & 5.21 & 1.14 & 1.01 & 3.74 & 0.57 \\
        \midrule
        \textbf{Avg (1-5)} & 3.15 & 4.96 & 0.96 & 1.11 & 2.66 & 0.46 \\
        \midrule
        6 & 3.28 & 4.94 & 0.80 & 1.01 & 2.77 & 0.44 \\
        7 & 3.29 & 4.57 & 0.87 & 1.02 & 2.05 & 0.32 \\
        8 & 3.10 & 6.12 & 1.23 & 1.08 & 2.46 & 0.51 \\
        9 & 3.38 & 4.26 & 0.54 & 1.08 & 1.79 & 0.28 \\
        10 & 3.33 & 5.71 & 0.88 & 1.19 & 3.16 & 0.45 \\
        \midrule
        \textbf{Avg (6-10)} & 3.28 & 5.12 & 0.86 & 1.08 & 2.45 & 0.40 \\
        \bottomrule
    \end{tabular}
    \label{tab:compare_3d_va}
\end{table}
\begin{figure}[h]
    \centering
	%trim={L,B,R,T}
	\includegraphics[trim=0cm 0cm 9cm 0cm,clip,width=0.95\linewidth]{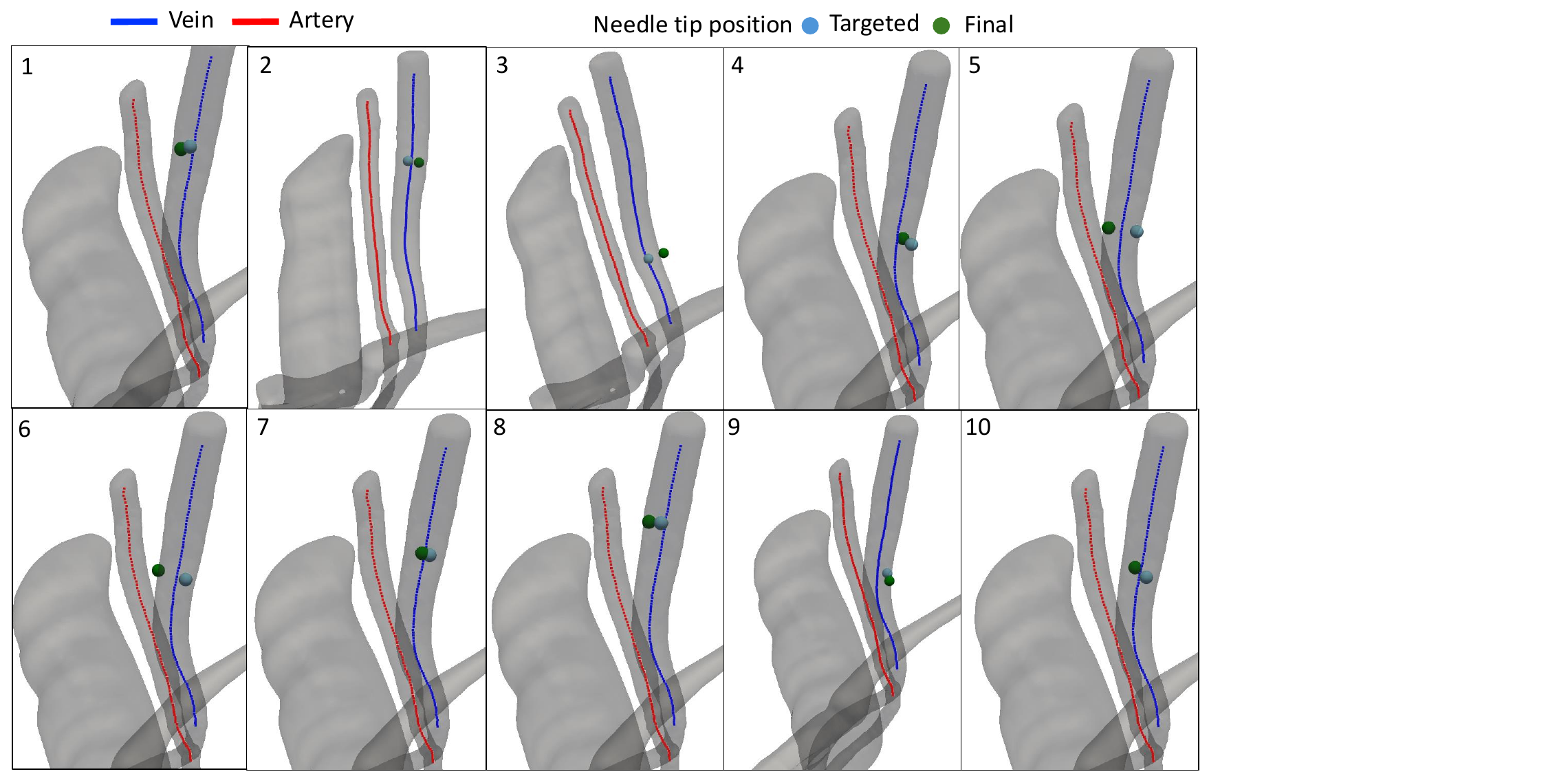}
    \caption{Comparing the 3D needle tip position (green sphere) with the targeted vein center-point (blue sphere)}
    \label{fig:needle_3d}
\end{figure}

% \begin{figure}[t]
%     \centering
% 	%trim={L,B,R,T}
% 	\includegraphics[trim=0cm 4.5cm 1cm 0cm,clip,width=0.8\linewidth]{figs/recon09}
%     \caption{Comparison of 3D reconstruction of centerline of vein and artery with its ground truth obtained from pre-operative CT scan}
%     \label{fig:recon_results}
% \end{figure}
\begin{figure}[!t]
    \centering
	%trim={L,B,R,T}
	\includegraphics[trim=0cm 6.2cm 6.2cm 0cm,clip,width=\linewidth]{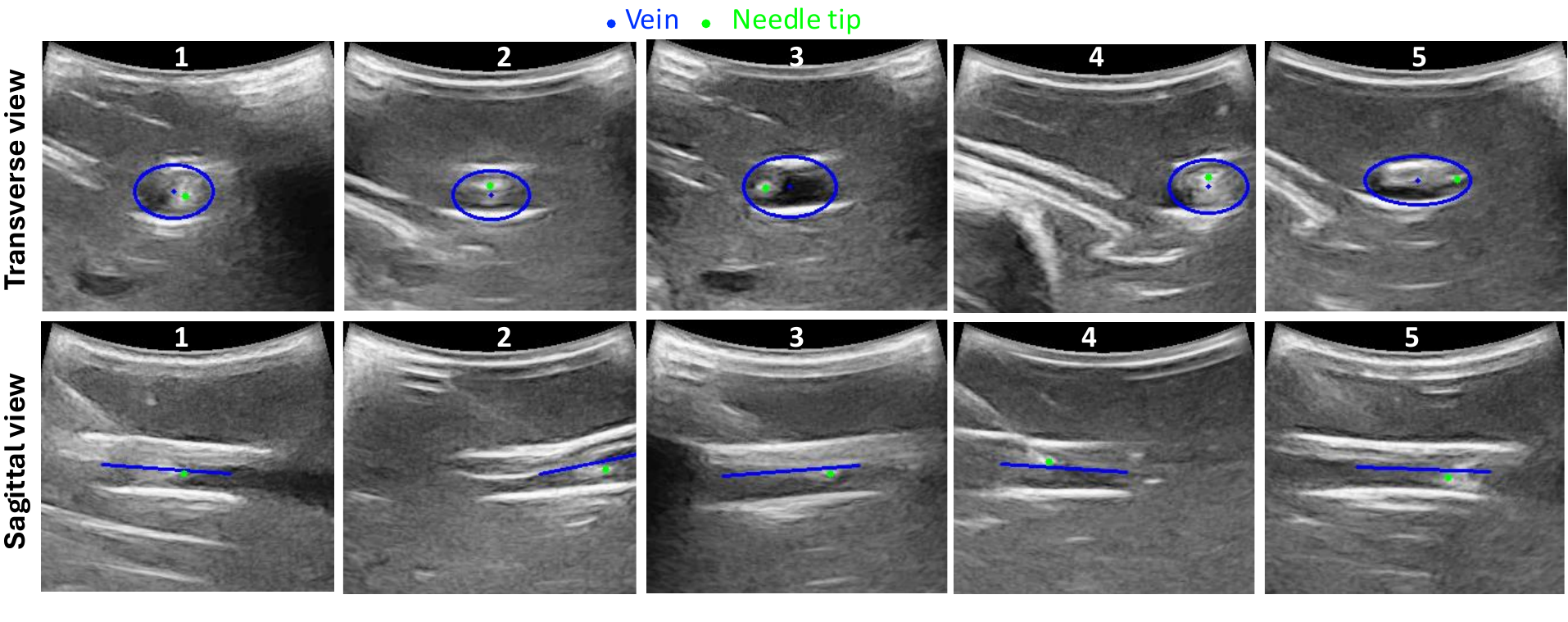}
	\includegraphics[trim=0cm 6.2cm 6.6cm 1cm,clip,width=\linewidth]{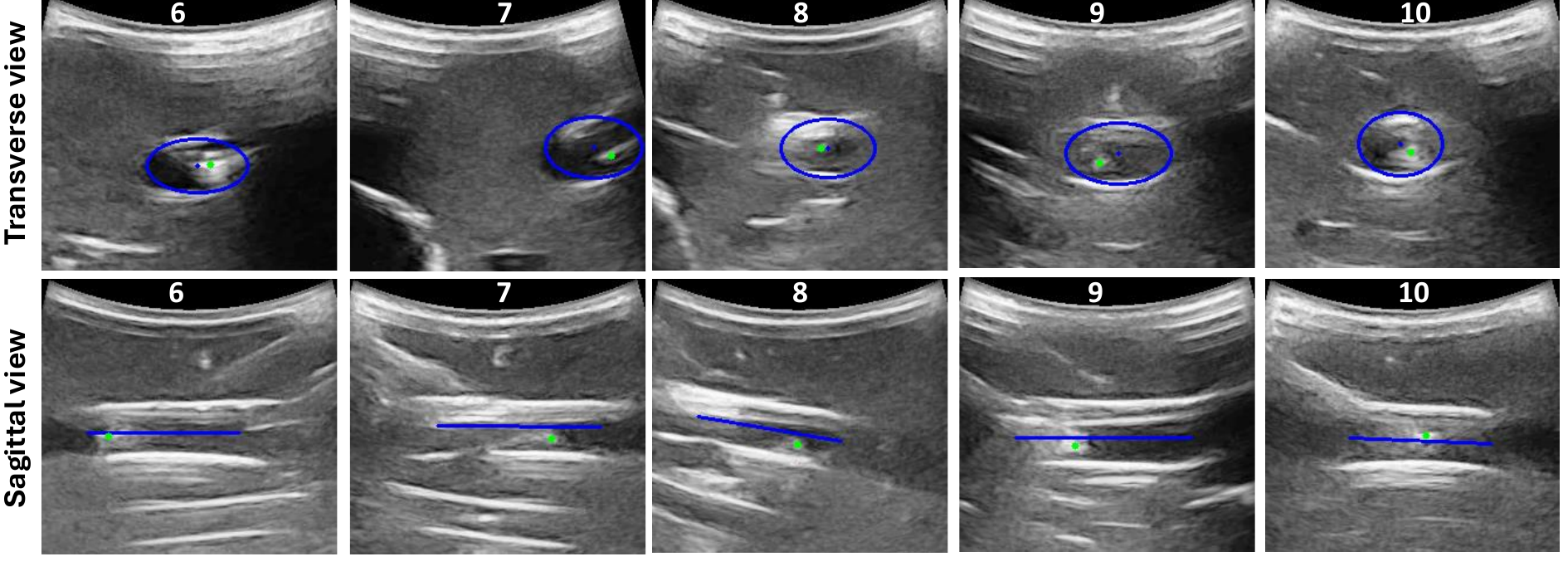}
    \caption{Expert's annotations on transverse and sagittal ultrasound views acquired with hand-held scanning after insertion}
    \label{fig:expert_view_markings}
\end{figure}
\subsection{Needle targeting accuracy}
The targeting accuracy of the system was evaluated by computing the error between the target and final needle tip position. The 3D positions of the target point and needle tip were calculated using Eq. \eqref{eq:pixel_to_3d} from the marked pixel locations in the sagittal view. After each experiment, an expert (co-author VP) marked the needle location and IJ vein (center point and boundary) in the ultrasound frame. This evaluation was performed first in the sagittal view after insertion. The probe was then removed, and manual handheld scanning collected transverse and sagittal views of the insertion region. Catheter insertion was performed to confirm vein access. Ultrasound videos from hand-held scanning and catheter insertion were later analyzed by the expert to verify safe insertion into the IJ vein.  

Fig. \ref{fig:needle_3d} shows the targeted 3D vein position and final needle tip location. Table \ref{tab:needle_3d_error} reports the corresponding errors under the `Evaluation via Robot' column. Fig. \ref{fig:expert_view_markings} shows expert annotations in selected transverse and sagittal frames from recorded ultrasound videos. The transverse view confirmed the needle tip remained inside the vein boundary, while the sagittal view re-confirmed that puncture was safe, with the tip remaining close to the centerline. Catheter insertion was also confirmed.  

Errors were computed as the distance between needle tip and vein center-point in the transverse view, and as the perpendicular distance to the vein centerline in the sagittal view. These distances were recorded in pixels and later converted to millimeters using the scaling factor described in Section \ref{sec:exp_testbed}. Table \ref{tab:needle_3d_error} summarizes these results under the `Evaluation via Expert' column.

\begin{table}[h!]
    \centering
    \caption{Error between the needle tip and vein center-point or center-line, evaluated with robot kinematics after insertion (Fig. \ref{fig:needle_3d}) and expert annotations (Fig. \ref{fig:expert_view_markings}). Insertion success is marked using '$\checkmark$' for safe needle insertion, bleeding of blue-color fluid (artery has red-color fluid), and catheter insertion.}
    \begin{tabular}{cccccccc}
    \toprule
     & \multicolumn{4}{c}{Error between needle tip and IJ vein's} &
    \multicolumn{3}{c}{Insertion Success}\\
    \cmidrule{2-3} \cmidrule{4-5} \cmidrule{6-8}
    ~ & \multicolumn{2}{c}{center-point ($mm$)} & \multicolumn{2}{c}{center-line ($mm$)} & Safe & Blue & Catheter \\
    \cmidrule{1-3} \cmidrule{4-5}
    % \midrule
    Evaluation via $\rightarrow$ & Robot & Expert & Robot & Expert & insertion & bleed & insertion\\
     % & kinematics & annotation & & & \\
    \midrule
    1 & 1.87 & 1.66 & \textcolor{blue}{0.39} & \textcolor{blue}{0.45} & \checkmark & \checkmark & \checkmark \\
    2 & 3.08 & 1.24 & 1.01 & 1.14 & \checkmark & \checkmark & \checkmark \\
    3 & 6.85 & 1.24 & \textcolor{blue}{0.90} & \textcolor{blue}{0.72} & \checkmark & \checkmark & \checkmark \\
    4 & 0.31 & 3.25 & \textcolor{blue}{0.47} & \textcolor{blue}{0.78} & \checkmark & \checkmark & \checkmark \\
    5 & 5.26 & 5.09 & \textcolor{blue}{0.32} & \textcolor{blue}{0.96} & \checkmark & \checkmark & \checkmark \\
    \midrule
    \textbf{Mean (1-5)} & \textbf{3.47} & \textbf{2.49} & \textbf{0.62} & \textbf{0.63} \\
    \textbf{S.D. (1-5)} & \textbf{2.61} &  \textbf{1.49} & \textbf{0.31} & \textbf{0.29} \\
    \midrule
    6 & 4.50 & 2.17 & 1.15 & \textcolor{blue}{0.46} & \checkmark & \checkmark & \checkmark \\
    7 & 1.27 & 2.49 & \textcolor{blue}{0.25} & 1.52 & \checkmark & \checkmark & \checkmark \\
    8 & 1.24 & 2.06 & \textcolor{blue}{0.80} & 1.85 & \checkmark & \checkmark & \checkmark \\
    9 & 2.19 & 1.57 & 1.38 & 1.32 & \checkmark & \checkmark & \checkmark \\
    10 & 2.18 & 2.01 & \textcolor{blue}{0.31} & \textcolor{blue}{0.69} & \checkmark & \checkmark & \checkmark \\
    \midrule
    \textbf{Mean (6-10)} & \textbf{2.28} & \textbf{2.06} & \textbf{0.78} & \textbf{1.17} \\
    \textbf{S.D. (6-10)} & \textbf{1.33} & \textbf{0.29} & \textbf{0.50} & \textbf{0.52} \\
    \bottomrule
    \end{tabular}
    \label{tab:needle_3d_error}
\end{table}
% During this evaluation, expert also marked the vein boundary using ellipses and needle tip in one of the axial US view in the recorded video. For the sagittal US view, centerline of the vein and needle tip has been marked. These evaluations by expert are shown in Fig. \ref{fig:expert_view_markings} for each of the experiment. It can be noted from transverse US views that for all the experiments the needle tip has been detected inside the vein. Further, sagittal view confirm that needle did not puncture through the vein.
% \subsubsection{Quantitative assessment}
% To quantitatively assess the performance of the proposed system in these experiments, we measured the error as the distance between the needle tip and the targeted point of IJ Vein. For evaluating the sagital frames after insertion while probe is still attached to robot, we  
% We also measured the maximum and minimum vein radius using the 3D Slicer VMTK module, obtaining values of $6.16$ $mm$ and $3.24$ $mm$, respectively.

\section{Discussion}
%\textbf{THE GENERAL INTENDED OUTLINE OF THIS SECTION WAS UNCLEAR. AN OPENING PARAGRAPH TO HELP PUT THE CONTENTS IN CONTEXT WAS MISSING AND SOME REORGANIZATION WAS NECESSARY. PLUS, WHAT ARE THE MAIN TAKEAWAY POINTS? I SUMMARIZED SOME OF THEM FROM MY PERSPECTIVE. SOME STATEMENTS SEEMED TO JUST RESTATE RESULTS THAT WERE ALREADY PRESENTED (WHICH IS NOT THE PURPOSE OF A DISCUSSION SECTION, SO I REMOVED THEM), FOLLOWED BY REASONS FOR POOR PERFORMANCE, WITHOUT FIRST TELLING US WHAT IS POSITIVE ABOUT THIS STUDY. SUGGESTION:}

% The work presented herein is the first to perform autonomous CLP with a depth camera, wireless ultrasound probe, and robotic force-controlled motion to identify the vein or artery. 
We presented phantom experiments to evaluate vessel centerline reconstruction and needle targeting accuracy. The proposed pipeline and validation serve as a foundation for a clinically deployable system. Performance was generally acceptable, with  three aspects of less-than-ideal performance  explained below.  

First, the reconstructed vein showed higher mean$\pm$S.D. error (Table \ref{tab:compare_3d_va}) due to two factors: (1) the  compressibility of the vein under force-controlled scanning, which was not accounted for in the CT ground truth, and (2) use of a convex rather than linear probe, which is typically preferred for CVC. Nonetheless, combined vein-artery error (2.15 mm) indicates the 2D-ultrasound-to-3D calibration preserved spatial accuracy, which is essential for needle entry planning.  

Second, needle targeting accuracy (Fig. \ref{fig:needle_3d}) showed tip alignment near the IJ vein centerline, with average errors of $0.62$ mm and $0.78$ mm for Exp. $1$–$5$ and $6$–$10$ (Table \ref{tab:needle_3d_error}). These values fall within the 1 mm threshold used in the controller (Section \ref{sec:needle_guidance}). Slightly higher errors arose in Exp. 2, 6 and 9 due to bending of the 150 mm long needle, yet reported values remained within the vein radius ($3.24$–$6.16$ mm).  

Third, for expert-evaluated manual scans, mean$\pm$SD errors from the center-point were $2.49 \pm 1.49$ mm and $2.06 \pm 0.29$ mm for Exp. $1$–$5$ and $6$–$10$, respectively, with largest errors in Exp. 4–5 from unavoidable needle tilting. The mean$\pm$SD errors from the centerline were $0.63 \pm 0.29$ mm and $1.17 \pm 0.52$ mm, for Exp. $1$–$5$ and $6$–$10$, respectively, increasing with induced trajectory noise but decreasing in Exp. 10 due to unpredictable bending. { Besides needle deflection, part of the error can be attributed to ultrasound calibration inaccuracies ($1.64 \pm 0.84$ mm). Nonetheless, across all experiments, mean needle tip error relative to the vessel centerline was $0.90 \pm 0.42$ mm, within the vein radius, confirming effective insertion.} For reference, Chen \textit{et al.} \cite{chen2021ultrasound} reported $0.9 \pm 0.29$ mm accuracy in a pork phantom.  

%{ \textbf{Limitations:}} 
{Three limitations should be addressed in future work. First, reducing the footprint of needle mechanism would improve positioning flexibility, which is particularly important for obese patients or those with shorter necks where sagittal access is not possible. Second, independent probe–needle motion could allow steeper insertion angles, better aligning with clinical preferences such as the lateral short-axis insertion emphasized in \cite{rossi2014percutaneous}. Finally, manual marking of the needle tip in ultrasound images currently limits real-time use; a future pipeline will integrate automated needle tip detection and tracking to improve autonomy and reduce operator dependence.}

{ %\textbf{
Future work %:} We 
will use anatomically representative phantoms \cite{al2025robotic} or fresh cadavers to  validate our system under more realistic physiological conditions. Our immediate next step would be validation of insertion zone selection on human subjects (without insertion). Beyond validation, we also aim to incorporate real-time tissue–needle interaction models and compensation strategies into our
framework.}

\section{Conclusion}
We have shown that the effective integration of intra-operative planning, anatomical identification, and needle guidance could enable the end-to-end robotic execution of central venous
catheterization. To the best of our knowledge, this is the first time depth camera feedback has been used to identify the scanning region, thereby eliminating the need to rely on medical experts or intra-operative CT/MRI scans. Another novelty of this work is the use of force-guided robotic motion to identify the vein and artery based on their compressibility, thereby reducing the reliance on complex ultrasound imaging feedback (multi-class segmentation or Doppler ultrasound). The success of this system can be observed from the successful needle placement with an error below or close to $1$ $mm$.

\bmhead{Acknowledgements}
This work is partially supported by the Agency for Healthcare Research and Quality (AHRQ) under Grant No. 7R18HS029124-03 (previously 5R18HS029124-03), the National Science Foundation (NSF) CAREER Award under Grant No. 2144348,
the NSF Future of Work at the Human-Technology Frontier (FW-HTF) Award No. 2222716, the NSF Alan T. Waterman Award No. 2431810, and the Malone Center for Engineering in Healthcare Postdoctoral Fellowship.

\bibliography{sn-bibliography}% common bib file
%% if required, the content of .bbl file can be included here once bbl is generated
%%\input sn-article.bbl

\end{document}